\documentclass[runningheads]{llncs}
\usepackage{graphicx}
\usepackage{amsmath,amssymb} 
\usepackage{color}
\usepackage{subcaption}
\usepackage{url}
\usepackage{caption}
\usepackage{tabularx}
\usepackage{multirow}
\usepackage{url}
\usepackage{booktabs}
\usepackage{comment}

\graphicspath{{Images/}}

\newcolumntype{C}{>{\centering\arraybackslash} X}
\newcolumntype{R}{>{\raggedleft\arraybackslash}X}%

\usepackage{mathtools}
\DeclarePairedDelimiter\floor{\lfloor}{\rfloor}

\newcommand{\pname}[1]{{{LAPRAN}}{#1}}

\begin{document}	
	\title{\pname{}: A Scalable Laplacian Pyramid Reconstructive Adversarial Network for Flexible Compressive Sensing Reconstruction} 
	
	\titlerunning{\pname{}: A Laplacian Pyramid Reconstructive Adversarial Network}
	
	\author{Kai XU\inst{1}\orcidID{0000-0001-8122-1419} \and
		Zhikang Zhang\inst{1}\orcidID{0000-0002-3024-1644} \and
		Fengbo Ren\inst{1}\orcidID{0000-0002-6509-8753}}
	%
	
	\authorrunning{K. XU et al.}
	%
	
	\institute{Arizona State University, Tempe AZ 85281, USA \\
		\email{ \{kaixu,zzhan362,renfengbo\}@asu.edu}
	}
	
	\maketitle
	
	\begin{abstract}
		This paper addresses the single-image compressive sensing (CS) and reconstruction problem. We propose a scalable Laplacian pyramid reconstructive adversarial network (\pname{}) that enables high-fidelity, flexible and fast CS images reconstruction. \pname{} progressively reconstructs an image following the concept of Laplacian pyramid through multiple stages of reconstructive adversarial networks (RANs). At each pyramid level, CS measurements are fused with a contextual latent vector to generate a high-frequency image residual. Consequently, \pname{} can produce hierarchies of reconstructed images and each with an incremental resolution and improved quality. The scalable pyramid structure of \pname{} enables high-fidelity CS reconstruction with a flexible resolution that is adaptive to a wide range of compression ratios (CRs), which is infeasible with existing methods. Experimental results on multiple public datasets show that \pname{} offers an average 7.47dB and 5.98dB PSNR, and an average 57.93$\%$ and 33.20 $\%$ SSIM improvement compared to model-based and data-driven baselines, respectively. Code is available at https://github.com/PSCLab-ASU/LAPRAN-PyTorch.
		
	\end{abstract}

	\section{Introduction}
Compressive sensing (CS) is a transformative sampling technique that is more efficient than Nyquist Sampling. Rather than sampling at the Nyquist rate and then compressing the sampled data, CS aims to directly sense signals in a compressed form while retaining the necessary information for accurate reconstruction. The trade-off for the simplicity of encoding is the intricate reconstruction process. Conventional CS reconstruction algorithms are based on either convex optimization \cite{Stephen2011nesta,Becker2011template,Dong2014nlrcs,Chengbo2009tval3,Metzler2016dcs} or greedy/iterative methods \cite{Thomas2009IHT,Huggins2007bp,Tropp2007omp}. These methods suffer from three major drawbacks limiting their practical usage. First, the iterative nature renders these methods computational intensive and not suitable for hardware acceleration. Second, the widely adopted sparsity constraint assumes the given signal is sparse on a known basis. 
However, natural images do not have an exactly sparse representation on any known basis (DCT, wavelet and curvelet) \cite{Metzler2016dcs}. The strong dependency on the sparsity constraint becomes the performance limiting factor of conventional methods. Constructing over-complete dictionaries with deterministic atoms \cite{Kai2016odl,kai2017sensing} can only moderately relax the constraint, as the learned linear sparsity models are often shallow thus have limited impacts. Third, conventional methods have a rigid structure allowing for reconstruction at a fixed resolution only. The recovery quality cannot be guaranteed when the compression ratio (CR) needs to be compromised due to a limited communication bandwidth or storage space. A better solution is to reconstruct at a compromised resolution while keeping a satisfactory reconstruction signal-to-noise ratio (RSNR) rather than dropping the RSNR for a fixed resolution.

Deep neural networks (DNNs) have been explored recently for learning the inverse mapping of CS \cite{srnet2016he,srcnn2014dong,srnet2016kim,ReconNet2016Kulkarni}. The limitations of existing DNN-based approaches are twofold. First, the reconstruction results tend to be blurry because of the exclusive use of a Euclidean loss. Specifically, the recovery quality of DNN-based methods are usually no better than optimization-based methods when the CR is low, e.g., $CR<=10$. 
Second, similar to the optimization-based methods, the existing DNN-based methods all have rigid structures allowing for reconstruction at a fixed and non-adaptive resolution only. The reconstruction will simply fail when the CR is lower than a required threshold.

In this paper, we propose a scalable Laplacian pyramid reconstructive adversarial network (\pname{}) for flexible CS reconstruction that addresses all the problems mentioned above. \pname{} does not require sparsity as prior knowledge hence can be potentially used in a broader range of applications, especially where the exact signal sparsity model is unknown. When applied to image signals, \pname{} progressively reconstruct high-fidelity images following the concept of Laplacian pyramid through multiple stages of specialized reconstructive adversarial networks (RANs). At each pyramid level, CS measurements are fused with a low-dimensional contextual latent vector to generate a reconstructed image
with both higher resolution and reconstruction quality. The non-iterative and high-concurrency natures of \pname{} make it suitable for hardware acceleration. Furthermore, the scalable pyramid structure of \pname{} enables high-fidelity CS reconstruction with a flexible resolution that can be adaptive to a wide range of CRs.  One can easily add or remove RANs from \pname{} to reconstruct images at a higher or lower resolution when the CR becomes lower and higher, respectively. Therefore, a consistently superior recovery quality can be guaranteed across a wide range of CRs.

The contributions of this paper are summarized as follows:
\begin{itemize}
    \item We propose a novel architecture of the neural network model (\pname{}) that enables high-fidelity, flexible and fast CS reconstruction.
    \item We provide to fuse CS measurements with contextual latent vectors of low-resolution images in each pyramid level to enhance the CS recovery quality.
    \item We illustrate that the progressive learning and reconstruction strategy can mitigate the difficulty of the inverse mapping problem in CS. Such a strategy not only accelerates the training speed by confining the search space but also improves the recovery quality by eliminating the accumulation of errors.
\end{itemize}
	\section{Related Work}
CS reconstruction is inherently an under-determined problem. Prior knowledge, i.e., the structure of signals must be exploited to reduce the information loss after reconstruction. According to the way of applying prior knowledge, CS reconstruction methods can be grouped into three categories: 1) model-based methods, 2) data-driven methods, 3) hybrid methods.
\subsection{Model-based Reconstruction Methods}
Model-based CS reconstruction methods mostly rely on the sparsity prior. For example, basis pursuit (BP), least absolute shrinkage and selection operator (LASSO), and least angle regression (LARS) are all based on $\ell_1$ minimization. Other methods exploit other types of prior knowledge to improve the recovery performance. NLR-CS \cite{Dong2014nlrcs} proposes a non-local low-rank regularization to exploit the group sparsity of similar patches. TVAL3 \cite{Chengbo2009tval3} and EdgeCS \cite{Weihong2010edgecs} use a total variation (TV) regularizer to reconstruct sharper images by preserving edges or boundaries more accurately. D-AMP \cite{Metzler2016dcs} extends approximate message passing (AMP) to employ denoising algorithms for CS recovery. 
In general, model-based recovery methods suffer from limited reconstruction quality, especially at high CRs. Because images, though compressible, are not ideally sparse in any commonly used transform domains \cite{Metzler2016dcs}. Additional knowledge of the image structure is required to further improve the reconstruction quality. Furthermore, when the available number of CS measurements is lower than the theoretical lower bound, the model-based methods would simply fail the reconstruction.
\subsection{Data-driven Reconstruction Methods}
Instead of specifying prior knowledge explicitly, data-driven methods have been explored to learn the signal characteristics implicitly. Kuldeep \textit{et al.} and Ali \textit{et al.} propose ``ReconNet'' \cite{ReconNet2016Kulkarni} and ``DeepInverse'' \cite{DeepInverse2017Mousavi}, respectively. Both work aims to reconstruct image blocks from CS measurements via convolutional neural networks (CNNs). Experimental results prove that both models are highly robust to noise and able to recover visually better images than the model-based approaches. However, the major drawback of these methods is the exclusive use of the $\ell_2$ reconstruction loss for training. As the $\ell_2$ loss cannot reliably generate shape images, additional loss metrics must be introduced to further improve the reconstruction quality. In addition, the direct mapping from the low-dimensional measurement domain to the high-dimensional image domain is highly under-determined. The under-determined mapping problem becomes even more notorious as CR increases since the dimension gap between the two domains is enlarged accordingly.
\subsection{Hybrid Reconstruction Methods} \label{sec:hybrid}
Hybrid methods aim to incorporate the benefits of both model-based and data-driven methods. Such methods first utilize expert knowledge to set up a recovery algorithm and then learn additional knowledge from training data while preserving the model interpretability and performance bounds. Inspired by the denoising-based approximate message passing (D-AMP) algorithm, Chris $\mathit{et al.}$ propose a learned D-AMP (LDAMP) network for image CS reconstruction. The iterative D-AMP algorithm is unrolled and combined with a denoising convolutional neural network (DnCNN) that serves as the denoiser in each iteration. The major drawback of this method is its sophisticated and iterative structure prohibiting parallel training and efficient hardware acceleration. 

Inspired by the success of generative adversarial network (GAN) for image generation, Bora \textit{et al.} propose to use a pre-trained DCGAN \cite{dcgan2015radford} for CS reconstruction (CSGM) \cite{Bora2017csgm}. This approach finds a latent vector $\hat{z}$ that minimizes the objective $\|A G(z) - y\|^2$, where $G$, $A$ and $z$ is the generator, sensing matrix and CS measurements, respectively. The optimal reconstruction result is represented as $G(\hat{z})$. Differently, the propose \pname{} directly synthesize an image from CS measurements, which alleviates the exploration of an additional latent space. Although both approaches are GAN-based, they represent two fundamentally different CS reconstruction schemes. CSGM is a sparse synthesize model \cite{Candes2005Incomplete,Candes2006robust} that approximates an unknown signal as $x=G(z)$, where the sparse coefficient ($z$) is measured concurrently. While \pname{} is a cosparse analysis model \cite{cosparse2013Nam,analysis2011Candes} that directly synthesize an unknown signal $x$ from the corresponding CS measurements $y$ according to $x=G(y)$. Hence we call the proposed model reconstructive adversarial network (RAN) instead of GAN. RAN elegantly approximates the nature image distribution from CS measurement samples, bringing the detour in the synthesize model unnecessary. \pname{} finishes reconstruction in a single feedforward propagation. While multiple network propagations are executed to obtain the optimal sparse coefficient $\hat{z}$ in CSGM. Therefore, \pname{} has a lower computational complexity and faster reconstruction speed.

	\begin{figure}[h]
	\centering
	\includegraphics[width=\textwidth]{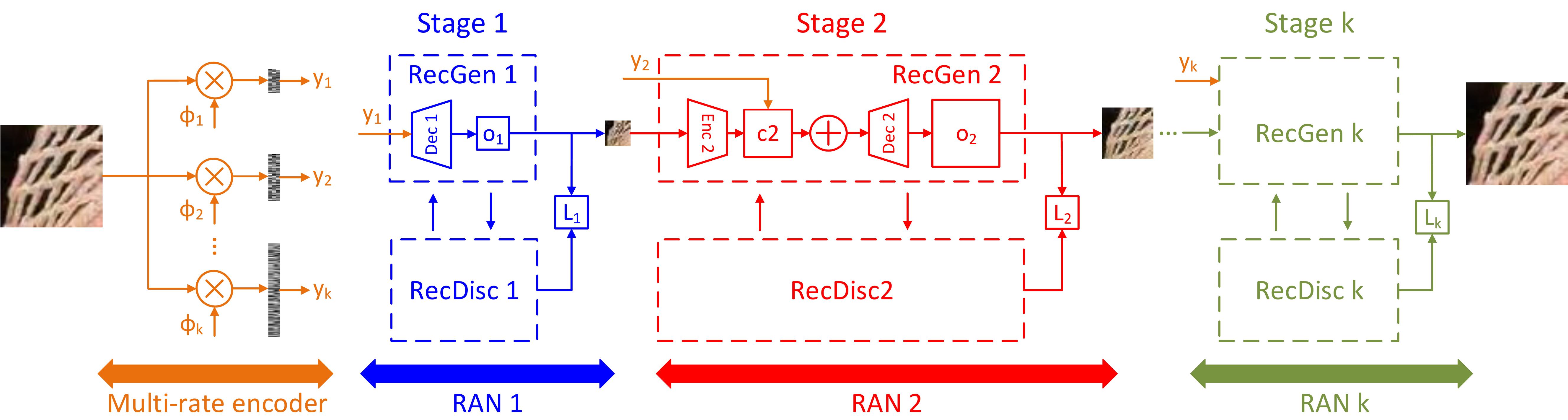}
	\caption{\textbf{Overall structure of the proposed \pname{}.} The CS measurement of a high-dimensional image is performed by a multi-rate random encoder. The \pname{} takes CS measurements as inputs and progressively reconstructs an original image in multiple hierarchies with incremental resolutions and recovery qualities. At each pyramid level, RAN generates an image residual, which is subsequently combined with an upscaled output from the previous level to form a higher-resolution output of the current level (upsampling and upscaling respectively refers to increasing the image resolution with and without new details added). The detailed structure of RAN is shown in Figure~\ref{fig:gen}.}
	\label{fig:overall}
\end{figure}
\section{Methodology}
The overall structure of the proposed CS system is shown in Figure~\ref{fig:overall}. It is composed of two functional units, a multi-rate random encoder for sampling and a \pname{} for reconstruction.  The multi-rate random encoder generates multiple CS measurements with different CRs from a single image. \pname{} takes the CS measurements as inputs and progressively reconstructs the original image in multiple hierarchies with incremental resolutions and recovery quality. In the first stage, RAN1 reconstructs a low-resolution thumbnail of the original image ($8 \times 8$). The following RANs at each stage fuses the low-resolution input generated by the previous stage with CS measurements to produce a reconstructed image upsampled by a factor of 2. Therefore, the resolution of the reconstructed image is progressively improved throughout the cascaded RANs. The proposed \pname{} architecture is highly scalable. One can concatenate more RANs (just like ``LEGO'' blocks) to gradually increase the resolution of the reconstructed image. Each building block of \pname{} is detailed below. Further details about the \pname{} architecture are provided in the supplementary materials.
\subsection{Multi-rate CS Encoder}
\begin{figure}[t]
	\centering
	\includegraphics[scale=0.27]{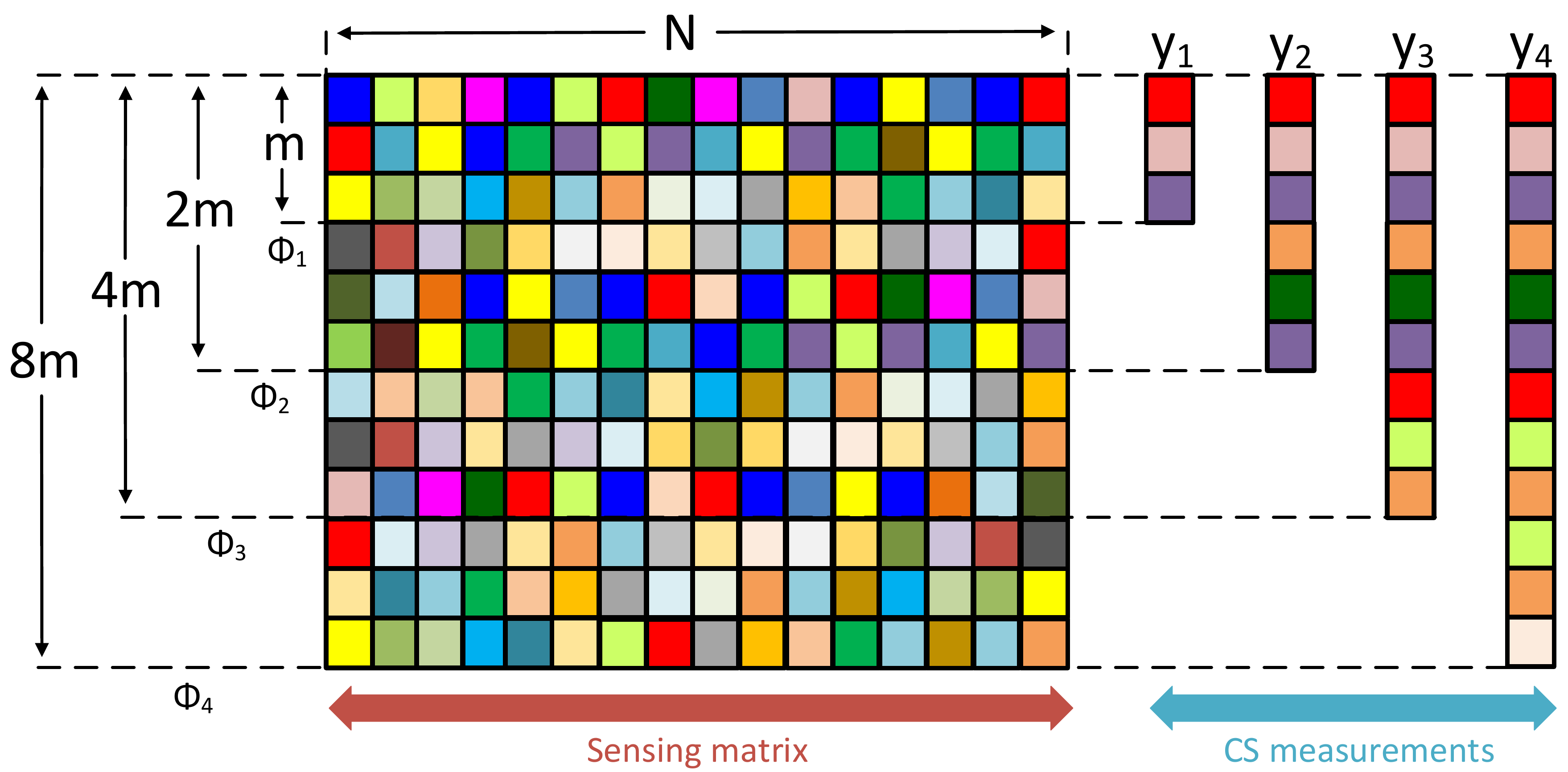}
	\caption{\textbf{Illustration of a sensing matrix for multi-rate CS}. The four sensing matrices $\mathbf{\Phi_1} \in \mathbb{R}^{m \times N}, \mathbf{\Phi_2} \in \mathbb{R}^{2m \times N}, \mathbf{\Phi_3} \in \mathbb{R}^{4m \times N}, \mathbf{\Phi_4} \in \mathbb{R}^{8m \times N}$ are used to generate the four CS measurements $\mathbf{\{y_1, y_2, y_3, y_4\}} \in \mathbb{R}^{\{m, 2m, 4m, 8m\}}$. $y_1, y_2, y_3, y_4$ is fed into RNN1 to RNN4 as the information source, respectively.}
	\label{fig:measurements}
\end{figure}

We propose a multi-rate random encoder for CS sampling. Given an input image, the encoder generates multiple CS measurements $\{\mathbf{y_1, \cdots, y_t}\}$ simultaneously, each has a different dimension. The generated measurements are fed into each stage of the RANs as input, i.e., $\{\mathbf{y_1, \cdots, y_k}\}$ is forward to \{RAN$1$, ..., RAN$k\}$, respectively. According to the rate-distortion theory \cite{Davisson1972rd}, the minimum bit-rate is positively related to the reconstruction quality, which indicates that the $i$-th RAN requires more information than all the previous RANs in order to improve the image resolution by adding finer details incrementally. The quantitative analysis of the number of measurements required for each RAN is as follows. Let $\mathbf{A}$ be an $m \times n$ sensing matrix that satisfies the restricted isometry property (RIP) of order $2k$, and the isometry constant is $\delta_{2k} \in (0,\frac{1}{2}]$. According to the CS theory \cite{Davenport2010random}, the lower bound of the number of CS measurements required for satisfying RIP is defined as: $m \geq C k \log(\frac{n}{k}),$ where $C=\frac{1}{2} \log(\sqrt{24}+1) \approx 0.28$.
In the CS image reconstruction problem, let the number of input measurements required by two adjacent RANs for accurately reconstructing a $N \times N$ image and a $2N \times 2N$ image is $m1$ and $m2$, respectively, we define the measurement increment ratio as $\beta=\frac{m2}{m1}$. If we assume the sparsity ratio $(\frac{k}{n})$ of the two images remains constant across the two adjacent RANs, then $\beta$ can be calculated as: 
\begin{align} \label{eq:2}
\beta = \frac{4k \times \log[(2N \times 2N)/4k]}{k \times \log[(N \times N)/k]} = 4.
\end{align}
Equation (\ref{eq:2}) indicates that the number of CS measurements (as well as CR) required for a former RAN should be at least 1/4 of a latter one in order to guarantee a satisfactory reconstruction performance. One should note that $\beta=4$ is the upper bound, lower $\beta$ values can be used to offer better reconstruction performance at the cost of collecting more CS measurements in early stages. In this work, we adopt $\beta=2$ to set a gradually increasing CR at different stages instead of using a unified CR. Since the dimension of a measurement vector equals to the number of rows in a sensing matrix, the $k$ sensing matrices in Figure~\ref{fig:overall} have the following dimensions: $\mathbf{\Phi_1} \in \mathbb{R}^{m \times N}, \mathbf{\Phi_2} \in \mathbb{R}^{\floor{\beta m} \times N}, \cdots, \mathbf{\Phi_k} \in \mathbb{R}^{\floor{\beta^{k-1}m} \times N}$. An example of the sensing matrix used for the multi-rate encoding of a 4-stage \pname{} is illustrated in Figure~\ref{fig:measurements}. The generated measurements $\mathbf{y_1} \in \mathbb{R}^m, \mathbf{y_2} \in \mathbb{R}^{2m}, \mathbf{y_3} \in \mathbb{R}^{4m}, \mathbf{y_4} \in \mathbb{R}^{8m}$ is used as the input to RAN1, RAN2, RAN3 and RAN4, respectively. With respect to a $k$-stage \pname{}, we only need to generate $\mathbf{y_t}$ for training. Since $\mathbf{y_i}$ is always a subset of $\mathbf{y_{i+1}}$, we can feed the first $\floor{\beta^{i-1}m}$ elements of $\mathbf{y_t}$ to the $i$-th stage in a backward fashion.

The proposed \pname{} enables CS reconstruction with a flexible resolution, which is not feasible with existing methods. When the number of CS measurements fail to meet the required threshold, the existing methods will fail to reconstruct with no room for maneuver. Alternatively, the proposed method can still reconstruct lower-resolution previews of the image with less detail in the case that the CS measurements are insufficient. The output of each RAN constitutes an image pyramid, providing the user with great flexibility in choosing the desired resolution of reconstructed images. 
\begin{figure}[t]
	\centering
	\includegraphics[width=\textwidth]{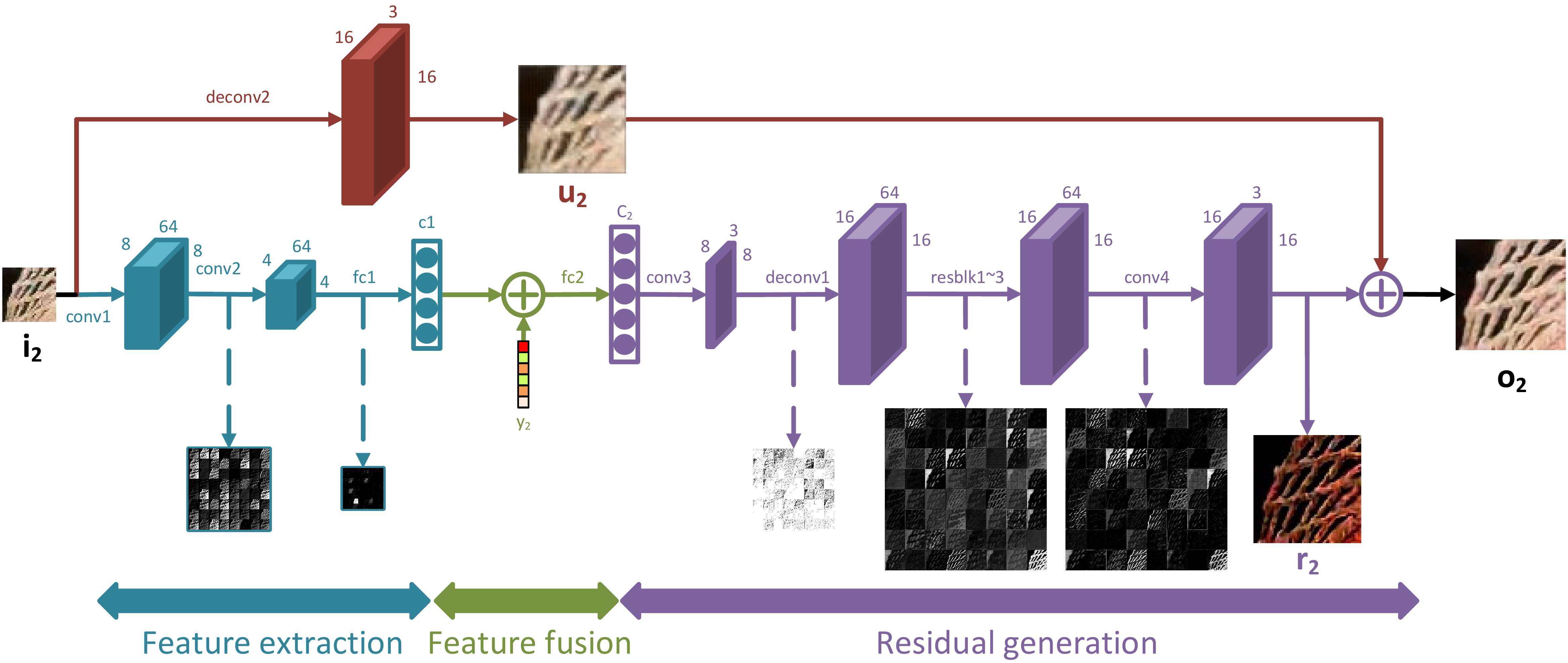}
	\caption{\textbf{The structure of RecGen2.} A low-resolution input image $\mathbf{i_2}$ is transformed into a high-frequency image residual $\mathbf{r_2}$ by an encoder-decoder network. A high-resolution output image is generated by adding the image residual to the upscaled input image. The dimension of each feature map is denoted in the figure. An example output of each convolutional layer is also shown.}
	\label{fig:gen}
\end{figure}
\subsection{RAN for CS Image Reconstruction} \label{sec:RAN}
We propose a RAN at each pyramid level to generate the reconstructed image with a fixed resolution. A RAN is composed of a reconstructive generator denoted as``RecGen", and a discriminator denoted as ``RecDisc." RecDisc follows the structure of DCGAN \cite{dcgan2015radford}, and the structure of RecGen is specially customized for reconstruction. Taking RecGen2 in the 2nd RNN stage as an example (see Figure~\ref{fig:gen}),  $\{\mathbf{i_2, r_2, u_2, o_2}\}$ is the contextual input from the previous stage, image residual, upscaled input, and output image, respectively. $\mathbf{y_2}$ is the input measurements generated by the multi-rate CS encoder. RecGen2 is composed of two branches: 1) the upper branch that generates an upscaled input image $\mathbf{u_2}$ via a deconvolutional neural network (deconv1); and 2) the lower branch that generates an image residual $\mathbf{r_2}$ to compensate for the artifacts introduced by the upper branch. Note that $\mathbf{u_2}$ is upscaled from a lower-resolution image, thus $\mathbf{u_2}$ lacks high-frequency components (see Figure~\ref{fig:gen}) and only provides a coarse approximation to the higher-resolution ground-truth image. It is the addition of the high-frequency residual $\mathbf{r_2}$ that recovers the entire frequency range of the image thus substantially improves the reconstruction quality \cite{Denton2015lapgan}.

The input $\mathbf{i_2}$ is treated as a low-resolution context for generating the residual image $\mathbf{r_2}$. We propose to first use an encoder to extract a contextual latent vector $\mathbf{c_1}$ to represent the low-resolution context $\mathbf{i_2}$. The encoder is composed of two convolutional layers and a fully-connected layer. To guarantee an equal contribution to the feature after fusion, the contextual latent vector $\mathbf{c_1}$ has the same dimension as the CS measurement $\mathbf{y_2}$. It should be noted that by increasing the dimension of $\mathbf{c_1}$, one can expect more image patterns coming from the contextual input appear in the final reconstruction, and vice versa. $\mathbf{c_1}$ is fused with the CS measurement $\mathbf{y_2}$ through concatenation (referred to as ``early fusion'' in \cite{Snoek2005fusion}) in a feature space. The fully-connected layer is used to transform the fused vector back to a feature map that has the same dimension as the contextual input $\mathbf{i_2}$.  A common practice of upscaling is to use an unpooling layer \cite{Zeiler2014unpool} or interpolation layer (bilinear, bicubic, or nearest neighbor). However, these methods are either non-invertible or non-trainable. Instead, we apply a deconvolutional layer deconv1 \cite{Zeiler2011deconv} to learn the upsampling of the fused feature map. We set up three residual blocks (resblk1$\sim$3) \cite{He2016DeepRL} to process the upsampled feature map to generate the image residual $\mathbf{r_2}$, which is later combined with $\mathbf{u_2}$ generated by the upper branch (deconv2) to form the final output image.  
\subsubsection{Learning from context.}
Instead of reconstructing the original image from CS measurements directly,  we propose to exploit the low-resolution context ($i_2$ in Figure~\ref{fig:gen}) to condition for reconstruction. The proposed conditional reconstruction scheme is fundamentally different from the conventional methods that solely rely on CS measurements. The reason is as follows.

Learning the inverse reconstructive mapping is a highly under-determined problem, hence notoriously tricky to solve. We need to accurately predict each pixel value in such an exceptionally high-dimensional space. All the existing data-driven methods directly search in such a vast space and try to establish a direct mapping from the low-dimensional CS measurements to the high-dimensional ground-truth.  The intricacy of the problem and the lack of additional constraints make the search process inefficient and untrustworthy. Differently, we delegate the low-resolution context to confine the sub-search space, i.e., the candidates that are far from the context in the search space will be obviated. Besides, the CS measurements supplement the necessary information needed for recovering the entire frequency spectrum of the image. The fusion of the context and CS measurements hence improve both convergence speed and recovery accuracy.

\subsubsection{Residual learning.}
In \pname{}, the RecGen of each RAN is similar to a segment of the ResNet in \cite{He2016DeepRL}. All the convolutional layers are followed by a spatial batch normalization (BN) layer \cite{Ioffe2015bn} and a ReLU except for the output layer. The output layer uses a Tanh activation function to ensure the output image has pixel values in the range of [0, 255]. The use of BN and normalized weight initialization \cite{LeCun1998prop} alleviates the problem of vanishing or exploding gradients hence improve both convergence accuracy and speed.

\subsection{Cascaded RANs for Flexible CS Reconstruction}
The existing DNN-based methods all have rigid structures allowing for reconstruction with a fixed CR and at a non-adaptive resolution only.  A new model must be retrained from scratch when a different CR is used in the encoding process. Inspired by the self-similarity based super resolution (SR) method \cite{Glasner2009sr,Zhen2014sr}, we propose a flexible CS reconstruction approach realized by dynamically cascading multiple RANs (see Figure~\ref{fig:overall}) at runtime. Upon training, each RAN corresponds to a specific resolution of the reconstructed image as well as an upper bound of the CR needed for accurate reconstruction. The thresholds of CR at different stages should be determined from experiments given a target accuracy metric. At runtime, depending on the CR of inputs, only the RANs with a higher CR threshold will be enabled for reconstruction. As a result, the proposed \pname{} can perform high-fidelity CS reconstruction with a flexible resolution that is adaptive to a wide range of CRs. This merit is particularly significant to the CS application scenarios, where the CR must be adaptive to the dynamic requirements of storage space or communication bandwidth. When the CR is compromised in such an application scenario, all the existing methods will fail the reconstruction, while the proposed \pname{} can still reconstruct an accurate preview of the image at a reduced resolution.

\begin{figure}[t]
    \centering
    \includegraphics[width=0.9\textwidth]{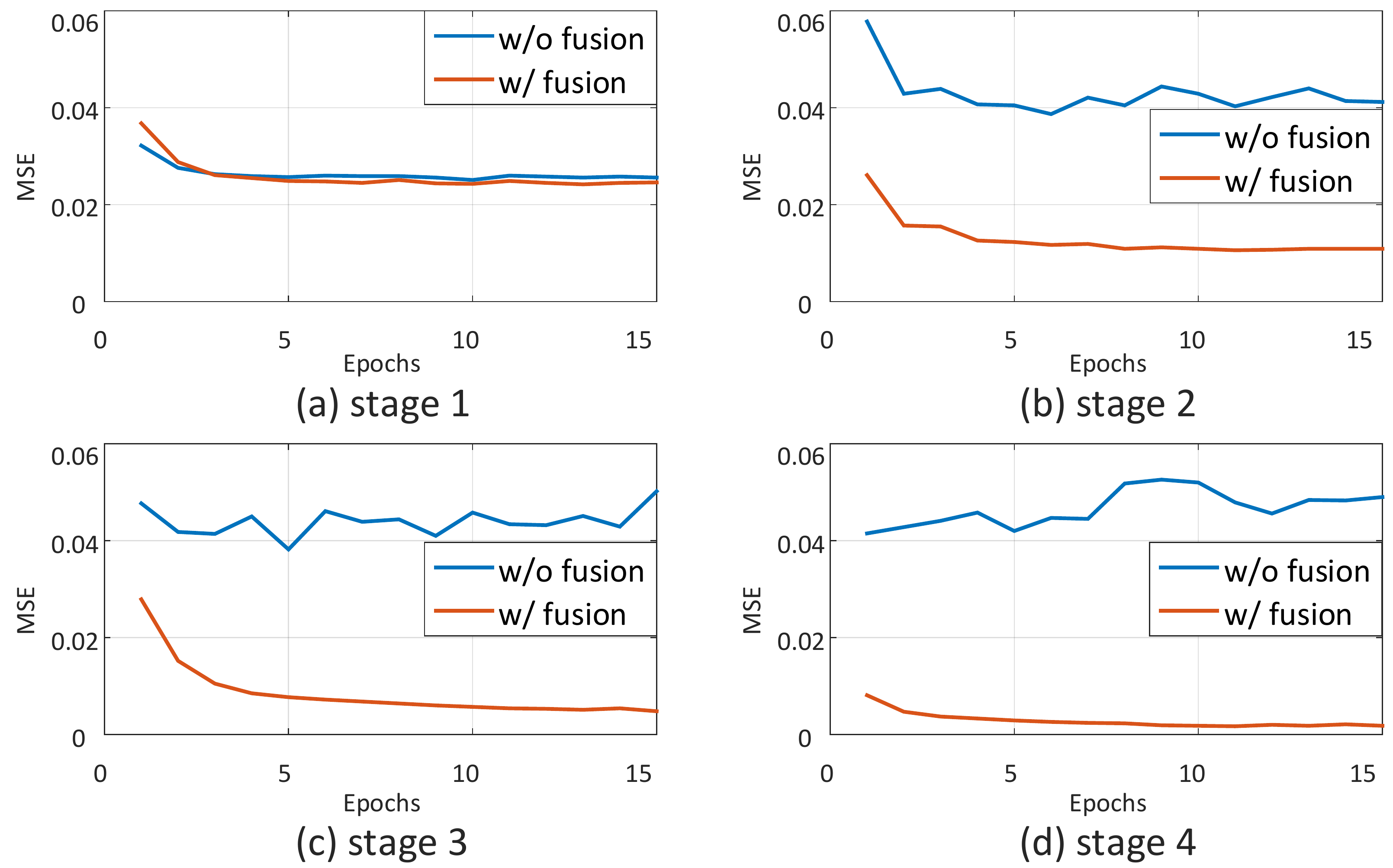}
    \caption{\textbf{Convergence analysis.} We compare the MSE test error using the CIFAR10 dataset at the CR of 10. The results without measurement fusion can be regarded as the performance of an SR approach. The MSE loss of the SR approach cannot be effectively reduced after stage 1 because of the lack of new information.}
    \label{fig:fusionCompare}
\end{figure}

Another advantage of the proposed \pname{} is that its hierarchical structure reduces the difficulty of training. CS reconstruction is a highly under-determined problem that has a humongous space for searching. Therefore, it is very challenging for a single network to approximate the inverse mapping accurately. Adopting a divide-and-conquer strategy, we propose to divide a highly under-determined problem into a series of lightly under-determined problems and conquer them in multiple hierarchies. As the dimensionality gap between the input and output in each sub-problem is significantly reduced, the difficulty for learning each mapping is much reduced compared to the original problem. Besides, since the hierarchical structure leverage a series of upsampling operations, error accumulation occurs at each stage. To alleviate such a problem, we define a loss function and perform back-propagation per stage independently. The training error is effectively reduced after each stage compared to the case that a single back-propagation is performed at the final output.
The injected CS measurements at each pyramid level are the key for CS reconstruction, which distinguishes the proposed approach from image SR methods. The SR models \cite{srnet2016he,srcnn2014dong,srnet2016kim,SRN2017lai} are responsible for inferring the high-frequency components non-existed in the input. From the frequency perspective, SR models should be adequately non-linear to compensate for the frequency gap, which inevitably results in complicated structures. Differently, the proposed approach incorporates new information provided by CS measurements into the reconstruction at each stage. The CS measurements supplement necessary information needed for recovering the entire frequency spectrum of an image, which is a powerful information source for learning visual representations. Consequently, both the resolution and the quality of the reconstructed images increase across different stages in the proposed approach. To illustrate this point, we compare \pname{} with a variant that has no fusion mechanism implemented at each stage (an SR counterpart). The comparison results are shown in Figure~\ref{fig:fusionCompare}. It is obvious that the reconstruction accuracy of the proposed \pname{} is consistently improved stage by stage, while the SR counterpart suffers from limited performance improvement.
\subsection{Reconstruction Loss}
We use a pixel-wise $\ell_2$ reconstruction loss and an adversarial loss for training. The $\ell_2$ loss finds an overall structure of a reconstructed image. The adversarial loss picks up a particular mode from the image distribution and generates a more authentic output \cite{Pathak2016context}. The overall loss function is defined as follows:
%
%
\begin{align}
\mathbf{z} &\sim \text{Enc}(\mathbf{z}|\mathbf{x_{l}}), \quad \mathbf{x_h} = G(\mathbf{y|z)}, \nonumber \\ 
\mathcal{L}_{adv}(G, D) &=     \mathbb{E}_{\mathbf{x_{h}}}[\log{D(\mathbf{x_{h}}|\mathbf{z})}] + \mathbb{E}_{\mathbf{y}}[\log(1-D(G(\mathbf{y}|\mathbf{z})))], \nonumber \\ 
\mathcal{L}_{euc} &= \mathbb{E}_{\mathbf{x_{h}}}[\|\mathbf{x_h - x_G}\|_2], \nonumber \\ 
\mathcal{L}_{total} &= \lambda_{adv}\mathcal{L}_{adv} + \lambda_{euc}\mathcal{L}_{euc},
\end{align}        
where $\mathbf{x_{l}}$, $\mathbf{x_{h}}$ and $\mathbf{x_G}, \mathbf{y}$ is the low-resolution input image, the high-resolution output image, the ground-truth image, and the CS measurement, respectively. The encoder function ($\text{Enc}$) maps a low-resolution input $\mathbf{x_{l}}$ to a distribution over a contextual latent vector $\mathbf{z}$.
\subsection{Training}
The training of each RAN is performed individually and sequentially. We start by training the first stage and the output is used as the input for the second stage. The training of all the subsequent stages is performed in such a sequential fashion. Motivated by the fact that the RANs in different stages share a similar structure but with different output dimensionality, we initialize the training of each stage with the pre-trained weights of the previous stage to take advantage of transfer learning. Such a training scheme is shown in experiments to be more stable and has faster convergence than those with static initialization (such as Gaussian or Xavier). Besides, the weight transfer between adjacent stages helps to tackle the notorious mode collapse problem in GAN since the pre-trained weights already cover the diversity existed in training images. It is recommended to leverage weight transfer to facilitate the training of the remaining RANs.
	\begin{table}[t]
	\centering
	\caption{\textbf{Summary of the major differences between the proposed and the reference methods.}}
	\label{tab:summary}	
	\begin{tabular}{c c c c c}
		\toprule
		Name     & Model/Data-driven & Iterative? & Reconstruction & Loss                      \\ 
		\midrule
		NLR-CS   & Model             & Yes        & Direct         & Group sparsity, low rank  \\
		TVAL3    & Model             & Yes        & Direct         & $\ell_2$, TV              \\ 
		D-AMP    & Model             & Yes        & Direct         & Denoising                 \\ 
		ReconNet & Data              & No         & Direct         & $\ell_2$                  \\ 
		LDAMP   & Hybrid            & Yes        & Direct         & Denoising                 \\ 
		CSGM     & Data              & No         & Direct         & $\ell_2$, Adversarial     \\ 
		\pname{}    & Data              & No         & Progressive    & $\ell_2$, Adversarial \\  
		\bottomrule
	\end{tabular}
\end{table}

\section{Experiments}
In this section, we evaluate the performance of the proposed method. We first describe the datasets used for training and testing. Then, the parameters used for training are provided. Finally, we compare our method with state-of-the-art CS reconstruction methods.
\subsection{Datasets and Training Setup}
We train and evaluate the proposed \pname{} with three widely used benchmarking datasets. The first two are MNIST and CIFAR10. The third dataset is made following the rule used in prior SR work \cite{srnet2016kim,SRN2017lai,Schulter2015forest}, which uses 91 images from Yang et al. \cite{Yang2010sr} and 200 images from the Berkeley Segmentation Dataset (BSD) \cite{Arbelaez2011detection}. The 291 images are augmented (rotation and flip) and cut into $228,688$ patches as the training data. Set5 \cite{Marco2012set5} and Set14 \cite{Zeyde2012set14} are pre-processed using the same method and used for testing. 

We implemented a 4-stage \pname{} for CS image reconstruction. We resize each training image to $64 \times 64$ and train the \pname{} with a batch size of 128 for 100 epochs with early stopping. We use Adam solver with a learning rate of $1 \times 10^{-4}$. The training takes roughly two days on a single NVidia Titan X GPU.

\subsection{Comparisons with State-of-the-art}
We compare the proposed \pname{} with six state-of-the-art CS reconstruction methods: NLR-CS \cite{Dong2014nlrcs}, TVAL3 \cite{Chengbo2009tval3}, BM3D-AMP (D-AMP with BM3D denoiser \cite{Dabov2007bm3damp}), ReconNet \cite{ReconNet2016Kulkarni}, CSGM \cite{Bora2017csgm}, and learned D-AMP\cite{Metzler2017ldamp}. All methods are summarized in Table~\ref{tab:summary}. Structural similarity (SSIM) and peak signal-to noise ratio (PSNR) are used as the performance metrics in the benchmarking.

\begin{figure}[t]
	\centering
	\begin{subfigure}[]{\textwidth}
		\centering
		\includegraphics[width=\textwidth]{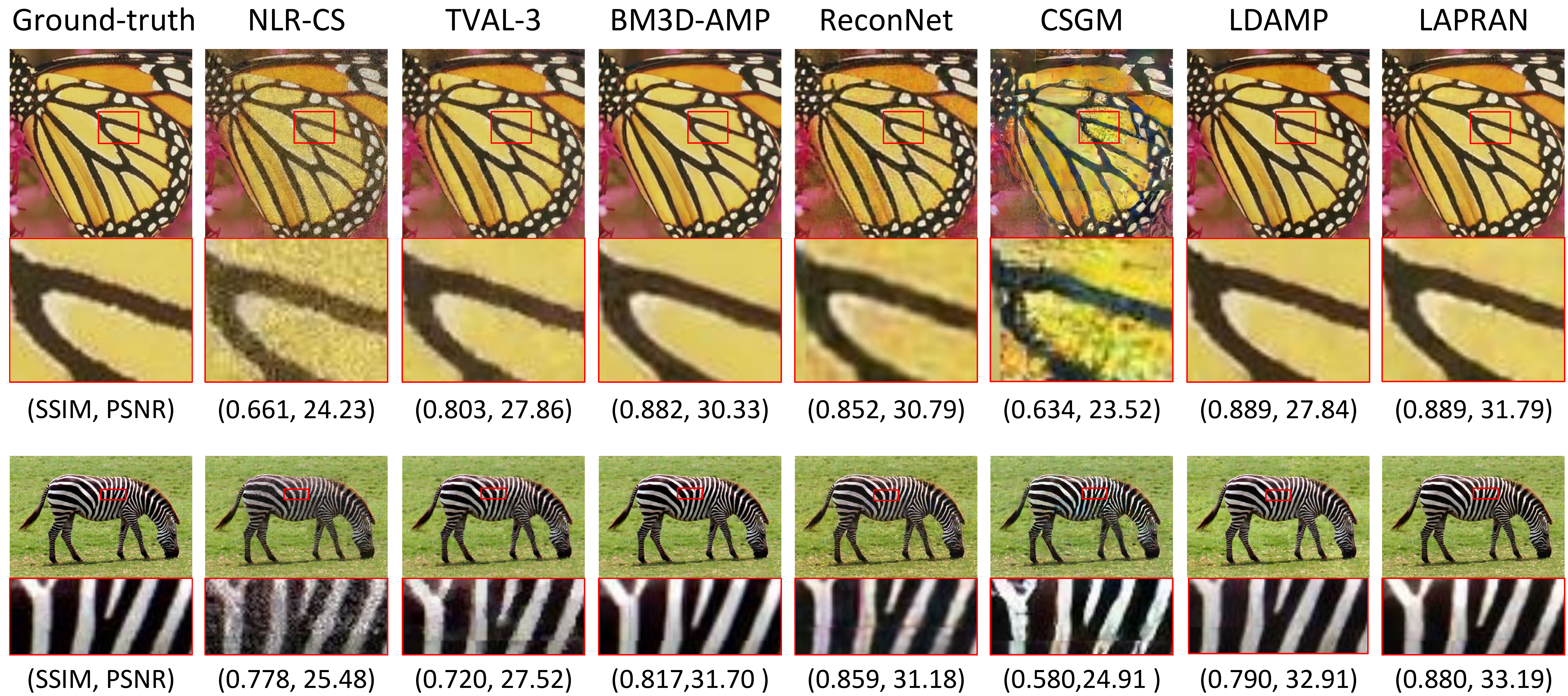} \\
		\caption{\textbf{CS reconstruction results at the CR of 5.}}
		\label{fig:cr5}
	\end{subfigure} \\
	\begin{subfigure}[]{\textwidth}
		\centering
		\includegraphics[width=\textwidth]{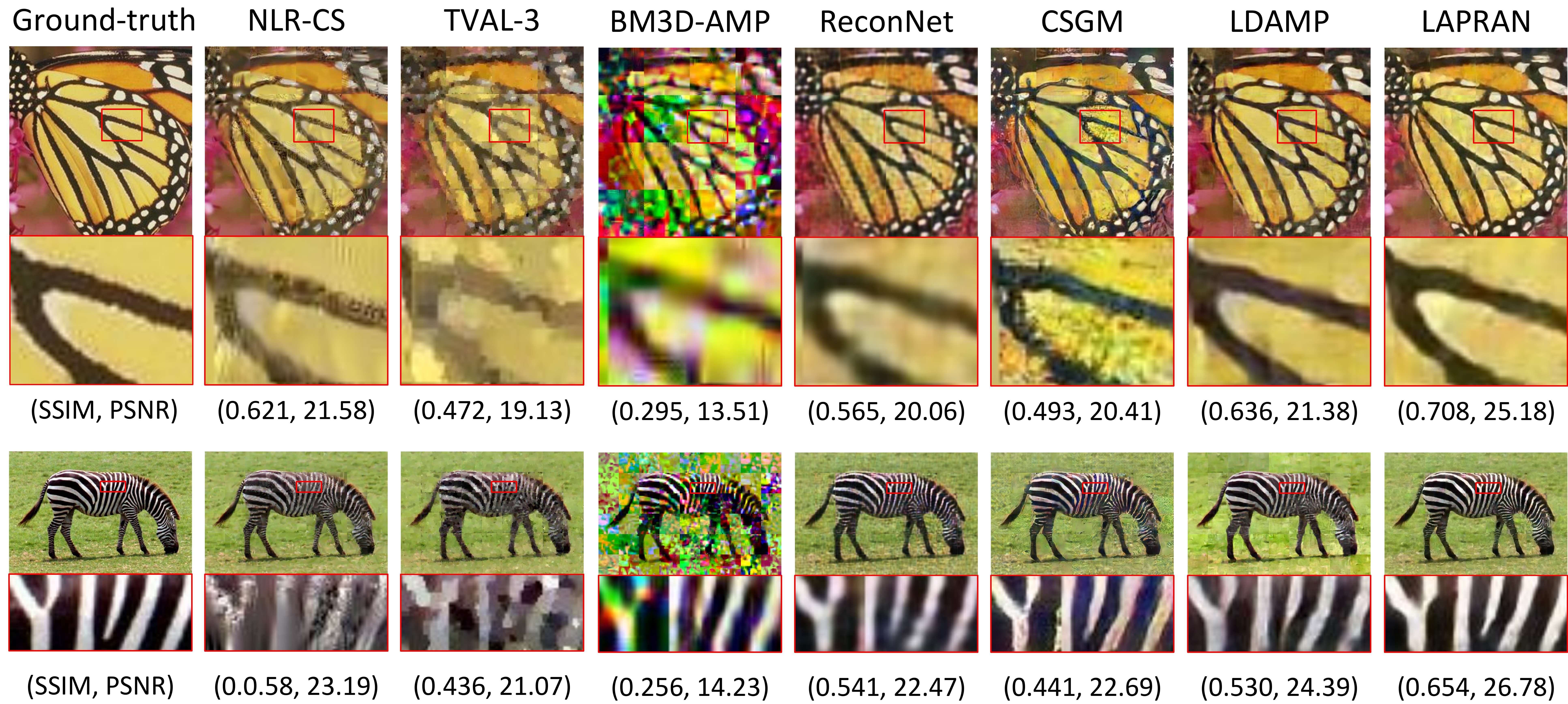} \\
		\caption{\textbf{CS reconstruction results at the CR of 20.}}
		\label{fig:cr20}
	\end{subfigure}
	\caption{\textbf{Visual comparison of butterfly (Set 5) and zebra (Set14) at the CR of 5 and 20, respectively. \pname{} preserves finer details.}}
\end{figure}

The quantitative comparison of reconstruction performance is shown in Table~\ref{tab:all}. The proposed \pname{} achieves the best recovery quality on all the testing datasets and at all CRs. Especially, the performance degradation of the \pname{} at large CRs ( $\geq$20) is well bounded. The main reasons are twofold. First, our approach adopts a progressive reconstruction strategy that greatly mitigates the difficulty of approximating the inverse mapping of CS. In contrast, CSGM tries to generate high-resolution images in a single step thus has a low reconstruction quality due to the difficulty in learning. Second, our approach utilizes a low-resolution image as input to guide the generation process at each stage, which helps to further reduce the search space of the under-determined problem by eliminating irrelevant candidates. The visual comparison of reconstructed images (at the CRs of 5 and 20) from Set 5 and Set 14 is shown in Figure~\ref{fig:cr5} and \ref{fig:cr20}, respectively. It is illustrated that our method can accurately reconstruct high-frequency details, such as the parallel lines, contained in the ground-truth image. In contrast, the reference methods produce noticeable artifacts and start to lose details at the CR of 20.
\begin{table}[htpb]
	\centering
	\caption{\textbf{Quantitative evaluation of state-of-the-art CS reconstruction methods.}}
	\label{tab:all}
	\begin{tabular}{|r|c|c|c|c|c|c|c|c|c|}
		\hline
		\multirow{2}{*}{ Algorithm } & \multirow{2}{*}{ $\enspace $ CR $\enspace $} & \multicolumn{2}{c|}{MNIST} & \multicolumn{2}{c|}{CIFAR10} & \multicolumn{2}{c|}{Set5} & \multicolumn{2}{c|}{Set14} \\ 
		\cline{3-10}
		& & SSIM & PSNR & SSIM & PSNR & SSIM & PSNR & SSIM & PSNR \\
		\hline
		\hline
		\multirow{1}{*}{NLR-CS} & \multirow{7}{*}{5}  
		  & 0.408 
		   & 24.85  
		    & 0.868 
		     & 37.91  
		      & 0.803 
		       & 30.42  
		        & 0.794 
		         & 29.42 \\
		D-AMP &                   
		 & {\color{blue} 0.983}  
		  & {\color{blue}37.78}     
		   & 0.968  
		    & 41.35  
		     & 0.852  
		      & {\color{blue}33.74}  
		       & {\color{blue}0.813} 
		        & {\color{blue}31.17} \\
		TVAL-3 &
		 & 0.934 
		  & 36.39 
		   & 0.847 
		    & 32.03  
		     & 0.812 
		      & 31.54  
		       & 0.727 
		        & 29.48 \\
		ReconNet &
		 & 0.911 
		  & 29.03          
		   & 0.871
		    & 32.55            
		     & 0.824 
		      & 31.78          
		       & 0.763 
		        & 29.70 \\
		CSGM &                     
		 & 0.748 
		  & 28.94  
		   & 0.788 
		    & 30.34  
		     & 0.619 
		      & 27.31  
		       & 0.575 
		        & 26.18 \\
		LDAMP &                     
		 & 0.797 
		  & 31.93           
		   & {\color{blue}0.971}
		    & {\color{blue}41.54}          
		     & {\color{blue}0.866}
		      & 32.26        
		       & 0.781 
		        & 30.07 \\
		\pname{} (ours) &                     
		 & {\color{red}0.993} 
		  & {\color{red}38.46}  
		   & {\color{red}0.978}
		    & {\color{red}42.39}  
		     & {\color{red}0.895} 
		      & {\color{red}34.79}  
		       & {\color{red}0.834}
		        & {\color{red}32.71} \\
		\hline
		\hline
		NLR-CS & \multirow{7}{*}{10} 
		 & 0.416
		  & 21.98  
		   & 0.840
		    & 33.39  
		     & 0.764 
		      &28.89  
		       & {\color{blue}0.716}
		        & 27.47 \\
		D-AMP &                     
		 & {\color{blue}0.963}
		  & {\color{blue}35.51}  
		   & 0.822
		    & 30.78  
		     & 0.743 
		      & 27.72  & 0.649
		       & 25.84 \\
		TVAL-3 &                     
		& 0.715
		 & 27.18  
		  & 0.746
		   & 29.21  
		    & 0.702
		     & 28.29  
		      & 0.615
		       & 26.65 \\
		ReconNet &                     
		& 0.868
		 & 28.98           
		  & 0.843
		   & 29.78          
		    & 0.779
		     & {\color{blue}29.53}         
		      & 0.704
		       & 27.45 \\
		CSGM &                     
		 & 0.589
		  & 27.49  
		   & 0.784
		    & 29.83  
		     & 0.560
		      & 25.82  
		       & 0.514
		        & 24.94 \\
		LDAMP &
		 & 0.446
		  & 22.40           
		   & {\color{blue}0.899}
		    & {\color{blue}34.56}          
		     & {\color{blue}0.796}
		      & 29.46          
		       & 0.687
		        & {\color{blue}27.70} \\
		\pname{} (ours) &                  
		 & {\color{red}0.990}
		  & {\color{red}38.38}  
		   & {\color{red}0.943}
		    & {\color{red}38.13}  
		     & {\color{red}0.849}
		      & {\color{red}32.53}  
		       & {\color{red}0.775}
		        & {\color{red}30.45} \\
		\hline
		\hline
		NLR-CS & \multirow{7}{*}{20} 
		 & 0.497
		  & 21.79  
		   & {\color{blue}0.820}
		    & {\color{blue}31.27}  
		     & 0.729
		      & 26.73  
		       & 0.621
		        & 24.88 \\
		D-AMP &                     
		 & 0.806
		  & {\color{blue}28.56}  
		   & 0.402
		    & 16.86  
		     & 0.413
		      & 16.72  
		       & 0.329
		        & 15.12 \\
		TVAL-3 &                     
		 & 0.494
		  & 21.00  
		   & 0.623
		    & 25.77  
		     & 0.583
		      & 25.18  
		       & 0.513
		        & 24.19 \\
		ReconNet &                     
		 & {\color{blue}0.898}
		  & 27.92          
		   & 0.806
		    & 29.08         
		     & {\color{blue}0.731}
		      & {\color{blue}27.07}       
		       & {\color{blue}0.623}
		        &{\color{blue}25.38} \\
		CSGM & 
		 & 0.512
		  & 27.54 
		   & 0.751
		    & 30.50  
		     & 0.526
		      & 25.04  
		       & 0.484
		        & 24.42 \\
		LDAMP &                     
		 & 0.346
		  & 17.01        
		   & 0.756
		    & 28.66         
		     & 0.689
		      & 27.00        
		       & 0.591
		        & 24.48 \\
		\pname{} (ours) &           
		 & {\color{red}0.985}
		  & {\color{red}37.02}  
		   & {\color{red}0.896}
		    & {\color{red}34.12}  
		     & {\color{red}0.801}
		      & {\color{red}30.08}  
		       & {\color{red}0.716}
		        & {\color{red}28.39} \\
		\hline
		\hline
		NLR-CS & \multirow{7}{*}{30} 
		 & 0.339
		  & 17.47  
		   & 0.703
		    & 27.26  
		     & 0.580
		      & 22.93 
		       & 0.581
		        & 22.93 \\
		D-AMP &                     
		 & 0.655
		  & 21.47  
		   & 0.183
		    & 10.62  
		     & 0.230
		      & 10.88  
		       & 0.136
		        & 9.31  \\
		TVAL-3 &                     
		 & 0.381
		  & 18.17  
		   & 0.560
		    & 24.01  
		     & 0.536
		      & 24.04  
		       & 0.471
		        & 23.20 \\
		ReconNet &                     
		 & {\color{blue}0.892}
		  & 25.46 
		   & {\color{blue}0.777}
		    & {\color{blue}29.32}           
		     & {\color{blue}0.623}
		      & {\color{blue}25.60}            
		       & {\color{blue}0.598}
		        &{\color{blue}24.59} \\
		CSGM &                     
		 & 0.661
		  & {\color{blue}27.47}  
		   & 0.730
		    & 27.73 
		     & 0.524
		      & 24.92  
		       & 0.464
		        & 23.97 \\
		LDAMP &                     
		 & 0.290
		  & 15.03       
		   & 0.632
		    & 25.57        
		     & 0.572
		      & 24.75         
		       & 0.510
		        & 22.74 \\
		\pname{} (ours) &
		& {\color{red}0.962}
		 & {\color{red}31.28}  
		  & {\color{red}0.840}
		   & {\color{red}31.47}  
		    & {\color{red}0.693}
		     & {\color{red}28.61}  
		      & {\color{red}0.668}
		       & {\color{red}27.09} \\
		\hline
	\end{tabular}
\end{table}
\begin{table*}[htpb]
	\centering
	\caption{\textbf{Runtime (seconds) for reconstructing a $64 \times 64$ image patch.} Unlike the model-based methods, the runtime of \pname{} is invariant to CR. \pname{} is slightly slower than ReconNet because of its large model capacity. CSGM and LDAMP are relatively slow due to their iterative nature.}
	\label{tab:time}
	\begin{tabularx}{\textwidth}{|C|C|C|C|C|C|}
		\hline
		Name   & Device  & CR=5     & CR=10    & CR=20    & CR=30    \\
		\hline
		\hline
		NLR-CS   & CPU  & 1.869e1  & 1.867e1  & 1.833e1  & 1.822e1  \\
		TVAL3    & CPU & 1.858e1  & 1.839e1  & 1.801e1  & 1.792e1  \\
		BM3D-AMP & CPU& 4.880e-1 & 4.213e-1 & 3.018e-1 & 2.409e-1 \\
		ReconNet & GPU& 2.005e-3 & 1.703e-3 & 1.524e-3 & 1.661e-3 \\
		CSGM     & GPU& 1.448e-1 & 1.125e-1 & 9.089e-2 & 8.592e-2 \\
		LDAMP   & GPU& 3.556e-1 & 2.600e-1 & 1.998e-1 & 1.784e-1 \\
		\pname{}   & GPU & 6.241e-3 & 6.384e-3 & 6.417e-3 & 6.008e-3 \\
		\hline
	\end{tabularx}
\end{table*}
\subsection{Reconstruction Speed}
We compare the runtime of each reconstruction method for reconstructing $64 \times 64$ image patches to benchmark reconstruction speed. For the optimization-based methods, GPU acceleration is ineffective due to their iterative nature. Thus, we use an Intel Xeon E5-2695 CPU to run the codes provided by the respective authors. For the DNN-based methods, we use a Nvidia GTX TitanX GPU to accelerate the reconstruction process. The average runtime for each method is shown in Table~\ref{tab:time}. The proposed \pname{} takes about 6ms to reconstruct $64 \times 64$ image patch, which is four orders of magnitude faster than NLR-CS and TVAL3, and two orders of magnitude faster than BM3D-AMP, LDAMP and CSGM. As illustrated in the section~\ref{sec:hybrid}, both LDAMP and CSGM are hybrid methods that require to solve a convex CS recovery problem. In each iteration, the DNN is propagated to provide a solution for a sub-problem. Therefore, multiple propagations are performed to obtain a single reconstruction, which explains why both LDAMP and CSGM are two orders of magnitude slower than \pname{}. In comparison with ReconNet, \pname{} sacrifices minor reconstruction speed for an apparent improvement in recovery quality (improves about 3$\sim$10dB PSNR). The proposed \pname{} is still sufficiently fast for performing real-time CS reconstruction. 

	\section{Conclusions}
%
%
In this paper, we present a scalable \pname{} for high-fidelity, flexible, and fast CS image reconstruction. The \pname{} consists of multiple stages of RANs that progressively reconstruct an image in multiple hierarchies. At each pyramid level, CS measurements are fused with a low-dimensional contextual latent vector to generate a high-frequency image residual, which is subsequently upsampled via a transposed CNN. The generated image residual is then added to a low-frequency image upscaled from the output of the previous level to form the final output of the current level with both higher resolution and reconstruction quality.  The hierarchical nature of the \pname{} is the key to enabling high-fidelity CS reconstruction with a flexible resolution that can be adaptive to a wide range of CRs. Each RAN in the \pname{} can be trained independently with weight transfer to achieve faster convergence and improved accuracy.  Leveraging the contextual input at each stage and the divide-and-conquer strategy in training are the keys to achieving excellent reconstruction performance.
\section{Acknowledgement}This work is supported by NSF grant IIS/CPS-1652038 and Google Faculty Research Award.


\begin{thebibliography}{8}
		\bibitem{Arbelaez2011detection}
		Arbelaez, P., Maire, M., Fowlkes, C., Malik, J.: Contour detection and
		hierarchical image segmentation. TPAMI  \textbf{33}(5),  898--916 (2011)
		
		\bibitem{Stephen2011nesta}
		Becker, S., Bobin, J., Candès, E.J.: Nesta: A fast and accurate first-order
		method for sparse recovery. SIAM Journal on Imaging Sciences  \textbf{4}(1),
		1--39 (2011)
		
		\bibitem{Becker2011template}
		Becker, S.R., Cand{\`e}s, E.J., Grant, M.C.: Templates for convex cone problems with applications to sparse signal recovery. Mathematical Programming	Computation  \textbf{3}(3),  165--218 (2011)
		
		\bibitem{Marco2012set5}
		Bevilacqua, M., Roumy, A., Guillemot, C., Alberi-Morel, M.L.: Low-complexity
		single-image super-resolution based on nonnegative neighbor embedding. In:
		BMVC, pp. 135.1--135.10 (2012)
		
		\bibitem{Thomas2009IHT}
		Blumensath, T., Davies, M.E.: Iterative hard thresholding for compressed
		sensing. Applied and Computational Harmonic Analysis  \textbf{27}(3),  265 --
		274 (2009)
		
		\bibitem{Bora2017csgm}
		Bora, A., Jalal, A., Price, E., Dimakis, A.G.: Compressed sensing using
		generative models. In: ICML, pp. 537--546 (2017)
		
		\bibitem{Candes2005Incomplete}
		{Candes}, E., {Romberg}, J., {Tao}, T.: {Stable Signal Recovery from Incomplete and Inaccurate Measurements}. Communications on Pure and Applied Mathematics \textbf{59}(8),  1207--1223 (2006)	
		
		\bibitem{Candes2006robust}
		Cand\`{e}s, E.J., Romberg, J., Tao, T.: Robust uncertainty principles: exact
		signal reconstruction from highly incomplete frequency information. TIT
		\textbf{52}(2),  489--509 (2006)
		
		\bibitem{analysis2011Candes}
		Candès, E.J., Eldar, Y.C., Needell, D., Randall, P.: Compressed sensing with
		coherent and redundant dictionaries. Applied and Computational Harmonic
		Analysis  \textbf{31}(1),  59 -- 73 (2011)
		
		\bibitem{Zhen2014sr}
		Cui, Z., Chang, H., Shan, S., Zhong, B., Chen, X.: Deep network cascade for
		image super-resolution. In: Fleet, D., Pajdla, T., Schiele, B., Tuytelaars, T. (eds.) ECCV 2014. LNCS, vol. 8693, pp. 49--64. Springer, Heidelberg (2014). \doi{10.1007/978-3-319-10602-1$\_$4}
		
		\bibitem{Dabov2007bm3damp}
		Dabov, K., Foi, A., Katkovnik, V., Egiazarian, K.: Image denoising by sparse
		3-d transform-domain collaborative filtering. TIP  \textbf{16}(8),  2080--2095 (2007)
		
		\bibitem{Davenport2010random}
		Davenport, M.A.: Random observations on random observations: Sparse signal
		acquisition and processing. Ph.D. thesis, Rice University (2010)
		
		\bibitem{Davisson1972rd}
		Davisson, L.: Rate distortion theory: A mathematical basis for data
		compression. TCOM  \textbf{20}(6),  1202--1202	(1972)
		
		\bibitem{Denton2015lapgan}
		Denton, E.L., Chintala, S., szlam, a., Fergus, R.: Deep generative image models using a laplacian pyramid of adversarial networks. In: NIPS, pp. 1486--1494 (2015)
		
		\bibitem{srnet2016he}
		Dong, C., Loy, C.C., He, K., Tang, X.: Image super-resolution using deep
		convolutional networks. TPAMI  \textbf{38}(2),  295--307 (2016)
		
		\bibitem{srcnn2014dong}
		Dong, C., Loy, C.C., He, K., Tang, X.: Learning a deep convolutional network
		for image super-resolution. In: Fleet, D., Pajdla, T., Schiele, B., Tuytelaars, T. (eds.) ECCV 2014. LNCS, vol. 8692, pp. 184--199. Springer, Heidelberg (2014). \doi{10.1007/978-3-319-10593-2$\_$13}
		
		\bibitem{Dong2014nlrcs}
		Dong, W., Shi, G., Li, X., Ma, Y., Huang, F.: Compressive sensing via nonlocal low-rank regularization. TIP  \textbf{23}(8),  3618--3632 (2014)
		
		\bibitem{Glasner2009sr}
		Glasner, D., Bagon, S., Irani, M.: Super-resolution from a single image. In:
		ICCV, pp. 349--356 (2009)
		
		\bibitem{He2016DeepRL}
		He, K., Zhang, X., Ren, S., Sun, J.: Deep residual learning for image
		recognition. In: CVPR, pp. 770--778 (2016)
		
		\bibitem{Huggins2007bp}
		Huggins, P.S., Zucker, S.W.: Greedy basis pursuit. TSP  \textbf{55}(7),  3760--3772 (2007)
		
		\bibitem{Ioffe2015bn}
		Ioffe, S., Szegedy, C.: Batch normalization: Accelerating deep network training by reducing internal covariate shift. In: ICML, pp. 448--456 (2015)
		
		\bibitem{srnet2016kim}
		Kim, J., Lee, J.K., Lee, K.M.: Accurate image super-resolution using very deep convolutional networks. In: CVPR, pp. 1646--1654 (2016)
		
		\bibitem{ReconNet2016Kulkarni}
		Kulkarni, K., Lohit, S., Turaga, P., Kerviche, R., Ashok, A.: Reconnet: Non-iterative reconstruction of images from compressively sensed
		measurements. In: CVPR, pp. 449--458 (2016)
		
		\bibitem{SRN2017lai}
		Lai, W.S., Huang, J.B., Ahuja, N., Yang, M.H.: Deep laplacian pyramid networks for fast and accurate super-resolution. In: CVPR, pp. 5835--5843 (2017)
		
		\bibitem{LeCun1998prop}
		LeCun, Y., Bottou, L., Orr, G.B., M\"{u}ller, K.R.: Efficient backprop.  In: Montavon G., Orr G.B., Müller KR. (eds.) Neural Networks: Tricks of the Trade. LNCS, vol 7700, pp. 9--48. Springer, Berlin, Heidelberg (2012). \doi{10.1007/978-3-642-35289-8$\_$3}
		
		\bibitem{Chengbo2009tval3}
		Li, C., Yin, W., , Zhang, Y.: An efficient augmented Lagrangian method with applications to total variation minimization. Computational Optimization and Applications \textbf{56}(3),  507--530 (2013)
		
		\bibitem{Metzler2016dcs}
		Metzler, C.A., Maleki, A., Baraniuk, R.G.: From denoising to compressed
		sensing. TIT  \textbf{62}(9),  5117--5144	(2016)
		
		\bibitem{Metzler2017ldamp}
		Metzler, C., Mousavi, A., Baraniuk, R.: Learned d-amp: Principled neural
		network based compressive image recovery. In: NIPS, pp. 1772--1783 (2017)
		
		\bibitem{DeepInverse2017Mousavi}
		Mousavi, A., Baraniuk, R.G.: Learning to invert: Signal recovery via deep
		convolutional networks. In: ICASSP, pp. 2272--2276 (2017)
		
		\bibitem{cosparse2013Nam}
		Nam, S., Davies, M., Elad, M., Gribonval, R.: The cosparse analysis model and
		algorithms. Applied and Computational Harmonic Analysis  \textbf{34}(1), 30--56 (2013)
		
		\bibitem{Pathak2016context}
		Pathak, D., Kr\"ahenb\"uhl, P., Donahue, J., Darrell, T., Efros, A.: Context
		encoders: Feature learning by inpainting. In:CVPR, pp. 2536--2544 (2016)
		
		\bibitem{dcgan2015radford}
		{Radford}, A., {Metz}, L., {Chintala}, S.: Unsupervised representation learning with deep convolutional generative adversarial networks. In: ICLR (2015)
		
		\bibitem{Schulter2015forest}
		Schulter, S., Leistner, C., Bischof, H.: Fast and accurate image upscaling with super-resolution forests. In: CVPR, pp. 3791--3799 (2015)
		
		\bibitem{Snoek2005fusion}
		Snoek, C.G.M., Worring, M., Smeulders, A.W.M.: Early versus late fusion in
		semantic video analysis. In: MM. pp. 399--402 (2005)
		
		\bibitem{Tropp2007omp}
		Tropp, J.A., Gilbert, A.C.: Signal recovery from random measurements via
		orthogonal matching pursuit. TIT \textbf{53}(12),  4655--4666 (2007)
		
		\bibitem{Weihong2010edgecs}
		Weihong~Guo, W.Y.: Edgecs: edge guided compressive sensing reconstruction.
		SPIE  \textbf{7744},  7744--7744 (2010)
		
		\bibitem{Kai2016odl}
		Xu, K., Li, Y., Ren, F.: An energy-efficient compressive sensing framework
		incorporating online dictionary learning for long-term wireless health
		monitoring. In: ICASSP, pp. 804--808 (2016)
		
		\bibitem{kai2017sensing}
		Xu, K., Li, Y., Ren, F.: A data-driven compressive sensing framework tailored
		for energy-efficient wearable sensing. In: ICASSP, pp. 861--865 (2017)
		
		\bibitem{Yang2010sr}
		Yang, J., Wright, J., Huang, T.S., Ma, Y.: Image super-resolution via sparse
		representation. TIP  \textbf{19}(11), 2861--2873 (2010)
		
		\bibitem{Zeiler2011deconv}
		Zeiler, M.D., Taylor, G.W., Fergus, R.: Adaptive deconvolutional networks for
		mid and high level feature learning. In: ICCV, pp. 1550--5499 (2011)
		
		\bibitem{Zeiler2014unpool}
		Zeiler, M.D., Fergus, R.: Visualizing and understanding convolutional networks. In: Fleet D., Pajdla T., Schiele B., Tuytelaars T. (eds.) ECCV 2014, LNCS, vol 8689, pp. 818--833. Springer, Cham (2014). \doi{10.1007/978-3-319-10590-1$\_$53}
		
		\bibitem{Zeyde2012set14}
		Zeyde, R., Elad, M., Protter, M.: On single image scale-up using
		sparse-representations. In: Curves and Surfaces, pp. 711--730 (2012)
	\end{thebibliography}
\end{document}